\documentclass[pmlr]{jmlr}%

\RequirePackage{graphicx}
 \usepackage{booktabs}
 \usepackage{dsfont}
\usepackage{longtable}%

\makeatletter
\def\set@curr@file#1{\def\@curr@file{#1}} %
\makeatother
\usepackage[load-configurations=version-1]{siunitx} %
\usepackage{multirow}
\usepackage{enumitem}
\usepackage{footnote}
\makesavenoteenv{table}
\makesavenoteenv{tabular}
\usepackage{adjustbox}
\usepackage{array}
\usepackage{booktabs}
\usepackage{subcaption}
\newcommand{\bftab}{\fontseries{b}\selectfont}

\theorembodyfont{\upshape}
\theoremheaderfont{\scshape}
\theorempostheader{:}
\theoremsep{\newline}

\jmlrvolume{298}
\jmlryear{2025}
\jmlrworkshop{Machine Learning for Healthcare}

\title[Stepwise Fine and Gray]{Stepwise Fine and Gray: Subject-Specific Variable Selection Shows When Hemodynamic Data Improves Prognostication of Comatose Post-Cardiac Arrest Patients}

\author{\Name{Xiaobin Shen} %
       \Email{xiaobins@andrew.cmu.edu}\\ 
       \addr Heinz College of Information Systems and Public Policy\\
       Carnegie Mellon University\\ \vspace{-0.8em}
       \\ 
       \Name{Jonathan Elmer} %
       \Email{elmerjp@upmc.edu}\\ 
       \addr Department of Emergency Medicine\\
       University of Pittsburgh\\ \vspace{-0.8em}
       \\
       \Name{George H. Chen} %
       \Email{georgechen@cmu.edu}\\ 
       \addr Heinz College of Information Systems and Public Policy\\
       Carnegie Mellon University\\
       }

\begin{document}

\maketitle
\vspace{-2em}
\begin{abstract}
  Prognostication for comatose post‐cardiac arrest patients is a critical challenge that directly impacts clinical decision-making in the ICU. Clinical information that informs prognostication is collected serially over time. Shortly after cardiac arrest, various time-invariant baseline features are collected (e.g., demographics, cardiac arrest characteristics). After ICU admission, additional features are gathered, including time-varying hemodynamic data (e.g., blood pressure, doses of vasopressor medications). We view these as two phases in which we collect new features. In this study, we propose a novel stepwise dynamic competing risks model that improves the prediction of neurological outcomes by automatically determining when to take advantage of time-invariant features (first phase) and time-varying features (second phase). A key finding is that it is not always beneficial to use all features (first and second phase) for prediction. Notably, our model finds patients for whom this second phase (time-varying hemodynamic) information is beneficial for prognostication and also \emph{when} this information is beneficial (as we collect more hemodynamic data for a patient over time, how important these data are for prognostication varies). %
  Our approach extends the standard Fine and Gray model to explicitly model the two phases and to incorporate neural networks to flexibly capture complex nonlinear feature relationships. %
  Evaluated on a retrospective cohort of 2,278 comatose post-arrest patients, our model demonstrates robust discriminative performance for the competing outcomes of awakening, withdrawal of life-sustaining therapy, and death despite maximal support. Subgroup analyses based on the motor component of the FOUR score reveal that patients with severe neurological dysfunction receive minimal additional prognostic benefit from hemodynamic data, whereas those with moderate-to-mild impairment derive significant incremental risk information. These findings underscore the potential of dynamic risk modeling for enhancing prognostication. Our approach generalizes to more than two phases in which new features are collected and could be used in other dynamic prediction tasks, where it may be helpful to know when and for whom newly collected features significantly improve prediction. %
  The source code implementing the proposed method is publicly available at \url{https://github.com/xiaobin-xs/Stepwise-Fine-and-Gray}.
\end{abstract}

\section{Introduction}

Prognostication in comatose survivors of cardiac arrest is a critical and challenging aspect of post-arrest care~\citep{rossetti2016neurological, geocadin2019standards, steinberg2024clinicians}. 
Among patients who remain in a coma and are admitted to the ICU, the leading cause of death is the withdrawal of life-sustaining therapy (WLST) based on physicians' perceived poor prognosis~\citep{may2019early}.
Guidelines focus primarily on predicting neurological outcomes, since death due to brain injury is more common than other irrecoverable organ failures. In particular, \emph{neuro}prognostication relies on a multimodal approach \citep{nolan2021european}, incorporating neurological exams, electroencephalogram (EEG), brain imaging, and blood biomarkers, as no single indicator reliably precludes recovery potential.
While the precision of neuroprognostic tools has improved in recent years (e.g., \citealt{westhall2016standardized}), %
accurate prognostication remains a substantial clinical challenge.

In addition to neuroprognostic testing, comatose survivors of cardiac arrest are critically ill and undergo close monitoring of many physiological processes in the intensive care unit. Continuous blood pressure monitoring is a core component of post-cardiac arrest intensive care, and the severity of low blood pressure (shock) has prognostic relevance~\citep{huang2017association, chi2022post}. 
According to post-resuscitation care guidelines~\citep{nolan2021european}, maintaining adequate mean arterial pressure (MAP) is essential, and hypotension (MAP$<$65 mmHg) should be avoided. Vasopressors are continuously administered medications that increase blood pressure to meet these treatment goals. 
Several studies have reported associations between early hypotension and poor neurological outcomes (e.g., \citealt{laurikkala2016mean, chiu2018impact}), while higher MAP levels have been linked to improved recovery ~\citep{beylin2013higher, roberts2019association}.
Compared to neuroprognostic tools like neurological exams and imaging, blood pressure signals are continuously available in the ICU. 
This provides a real-time window into cardiovascular stability, which may indirectly reflect cerebral perfusion and brain health. 
Thus, hemodynamic signals could complement other modalities in a multimodal neuroprognostic strategy. %

Machine learning approaches have recently been applied to neuroprognostication using multimodal predictors, including EEG and blood pressure (e.g., \citealt{zheng2021predicting, kim2023machine, hessulf2023predicting, kim2025system}). However, many existing models treat neuroprognostication as a time-invariant, binary classification task by contrasting good versus poor outcomes. This approach may introduce bias in the target labels, especially in the context of WLST. Patients who die following WLST are often labeled as having poor outcomes, despite the uncertainty of their true neurological trajectory had life-sustaining therapy been continued~\citep{10.1001/jamanetworkopen.2025.1714}. On the other hand, excluding these patients entirely would not only waste valuable data but could also introduce a sampling bias: if, for instance, the vast majority of these patients would have ended up dying despite maximal support, then excluding patients who died following WLST would bias the remaining dataset to have a larger proportion of patients with ``good'' outcomes (e.g., who are likely to awaken). %

Instead of using a binary classification approach, a more principled solution is to adopt a competing risks setup (see, e.g., Chapter 12 of the standard text by \citet{collett2023modelling}), a standard extension of survival analysis to model the time that will elapse until the earliest of multiple mutually exclusive outcomes (e.g., awakening from coma, death following WLST, death despite maximal support), while still allowing for censoring (e.g., a patient still being in a coma by the time data collection stops). Especially as we are in a setting where we collect more data on each patient over time, we specifically adopt a \emph{dynamic} competing risks setup, with predictions repeatedly made over time (see, e.g., Section 6.2 of \citet{chen2024introduction}). %

In this study, we propose a ``stepwise'' neural net extension to the standard competing risks method by \citet{fine1999proportional} to evaluate the role of hemodynamic data (including blood pressure and vasopressor dosages over time) in predicting neurological outcomes in comatose post-cardiac arrest patients. The stepwise nature could be described in terms of two phases, which correspond to the ordering in which features are commonly collected: in the first phase (shortly after cardiac arrest), we collect static features (e.g., demographics, cardiac arrest characteristics, neurological exams) and can already make predictions using these features. Later in the second phase (after ICU admission), we start collecting time-varying features (e.g., blood pressure trajectories, vasopressor usage), which we use to refine predictions from the already built first phase model as to come up with a second phase model. Our method learns a thresholding rule to decide whether to use the first or second phase model to make predictions, where the thresholding depends on the specific test patient as well as how much of the patient's data we have seen. In other words, we determine \emph{for whom} and \emph{when} time-varying hemodynamic features are particularly important for prediction over static baseline features. This is in contrast to other competing risks variable selection approaches based on boosting or lasso or certain selection criteria (like AIC and BIC) that select which features to use across all subjects and thus do not provide variable selection personalized at the individual subject level (e.g., \citealt{binder2009boosting, kuk2013model, tapak2015competing,fu2017penalized}). %

We apply our proposed stepwise Fine and Gray method on 2,278 comatose post-arrest patients. Our results suggest that our approach improves model performance (in terms of a competing risks version of the concordance index \citep{wolbers2014concordance,ishwaran2014random}) compared to using all features simultaneously.
We also provide individual-level dynamic risk profiles over time. Changes in predicted risk over time, quantified as the ratio of log subhazards, between phases serve as potential alerts for clinicians to reassess a patient's neurological status or to consider intervention. 
Additionally, subgroup analyses based on the motor component of the FOUR score~\citep{wijdicks2005validation} reveal that patients with moderate-to-mild early neurological impairment benefit most from continuous hemodynamic monitoring, as evidenced by the significant incremental prognostic contribution from time-varying blood pressure and vasopressor dosage features.
We believe our findings offer meaningful insights into the integration of continuous hemodynamic data into neuroprognostication workflows and may inform future clinical decision-making.

\subsection*{Generalizable Insights about Machine Learning in the Context of Healthcare}
From a methodological viewpoint, our learning framework illustrates how decomposing risk prediction into early static features and subsequent time-varying data can yield interpretable insights that could aid clinical decision-making. This approach is flexible and easily extends to arbitrarily many phases of newly collected features per subject, providing insight as to which features are significantly beneficial for prediction for which subjects and when.

From a decision support viewpoint, by quantifying the incremental prognostic contribution of dynamic measurements (the hemodynamic features in our case), our method demonstrates the value of continuous monitoring in refining risk estimates. Although blood pressure control is a core component of post-cardiac arrest care, blood pressure data is not typically incorporated into prognostication algorithms for these patients. Our analysis reveals that these dynamic hemodynamic data possess predictive power that could potentially complement other modalities such as EEG.

The framework and evaluation strategy presented here are not limited to prognostication. They can be adapted to other time-to-event prediction problems in healthcare, such as treatment response or disease progression, where distinguishing the impact of static versus dynamic factors (or other ways of splitting features into phases) could be beneficial.

\section{Background}\label{sec:background}

We first review the competing risks setup (Section~\ref{sec:setup}) followed by the classic \citet{fine1999proportional} competing risks model (Section~\ref{sec:fg-model}). For the latter, we go over why the Fine and Gray model can easily be interpreted, which is why we extend it in our proposed approach. Throughout the paper, for any positive integer $\ell$, we use $[\ell] \triangleq \{1,2,\ldots, \ell\}$ to denote the set of integers from 1 to $\ell$. Random variables are written as uppercase letters (e.g., $X$) whereas realizations of random variables, dummy variables, and constants are written as lowercase letters (e.g., $x$). For reference, we provide a list of all notations used in Appendix~\ref{app:notation}.

\subsection{Competing Risks Problem Setup}
\label{sec:setup}

Suppose that there are $m$ competing events that are mutually exhaustive. We now describe the statistical setup assumed for the training data and, separately, for test data.

\paragraph{Training data} For each training subject $i\in[n]$, we observe the tuple $(X_i,Y_i,D_i)$, where $X_i\in\mathcal{X}$ is the $i$-th subject's feature vector, $Y_i\in(0,\infty)$ is the $i$-th subject's ``observed time'' (to be defined momentarily), and $D_i\in\{0,1,\dots,m\}$ indicates which competing event is the earliest to happen for the $i$-th subject (the special value of 0 means that the censoring happens before any of the $m$ competing events). If $D_i=0$, then $Y_i$ is the censoring time. Otherwise (i.e., if $D_i\in[k]$), then $Y_i$ is the time until the earliest competing event happened.

Here, we could view the $i$-th subject as having their $Y_i$ and $D_i$ random variables generated based on the latent random variables $T_{i,1}, T_{i,2}, \ldots, T_{i,m}, C_i$, where $T_{i,k}$ (for $k \in [m]$) is the time of occurrence for event $k$ of the $i$-th subject, and $C_i$ is the censoring time for the $i$-th subject. We assume that conditioned on feature vector $X_i$, the event times $(T_{i,1}, T_{i,2}, \ldots, T_{i,m})$ are independent of the censoring time $C_i$. Then
\begin{align*}
Y_i &\triangleq \min\{ T_{i,1}, T_{i,2}, \ldots, T_{i,m}, C_i \}, \\
D_i &\triangleq
\begin{cases}
0 & \text{if } Y_i = C_i \text{ (censoring is the earliest to happen)}, \\
\arg\min_{k \in [m]} T_{i,k} & \text{otherwise.}
\end{cases}
\end{align*}

\paragraph{Test data and prediction}
At test time, we assume a slightly different statistical model. We denote a generic test subject's feature vector as $X$, and we use $T_k$ (for $k\in[m]$) to denote the time of occurrence of event $k$. In particular, the tuple $(X,T_1,\dots,T_m)$ has the same distribution as the tuple $(X_i,T_{i,1},\dots,T_{i,m})$ for any training subject $i\in[n]$. We are not interested in reasoning about whether the test subject will be censored (so that we do not define a censoring time for the test subject). We define the time until the earliest event happens $T$ and the event indicator $D$ as
\begin{equation*}
T \triangleq \min_{k\in[m]} T_k,
\qquad\text{and}\qquad
D \triangleq \arg\min_{k\in[m]} T_k.
\end{equation*}
For any observed test feature vector $X\!=\!x\in\!\mathcal{X}$, our goal is to predict the so-called \emph{cumulative incidence function} (CIF) for every event $k \in [m]$, defined for any time horizon $h>0$ as  
\begin{equation}
F_k(h|x) \triangleq \mathbb{P}(T\leq h, D=k \mid X=x),
\label{eq:cif}
\end{equation}
which can be interpreted as the probability of event $k$ occurring before time horizon $h$ conditioned on feature vector $x$.

\subsection{Fine and Gray Competing Risks Model}
\label{sec:fg-model}

We now review the classic \cite{fine1999proportional} competing risks model, which our proposed approach extends. The Fine and Gray model in turn extends the standard Cox proportional hazards model \citep{cox1972regression} to the competing risks setting. To do so, we work with the \emph{subdistribution hazard function} (abbreviated as \emph{subhazard}), defined for each event $k\in[m]$~as %
\begin{equation}
\lambda_k(h|x)
\triangleq
  \lim_{\Delta h \to 0}
    \frac{1}{\Delta h} 
    \mathbb{P} \Big(T\in[h, h + \Delta h], D = k ~\Big|~ \big(T \geq h \cup \underbrace{(T \leq h, D \neq k)}_{\text{``unnatural'' condition}}\big), X = x \Big)
\label{eq:subhazard}
\end{equation}
Importantly, in the conditional probability above, there is an ``unnatural'' condition. First off, it is natural to consider a subject to still be at risk at time $h$ if they have not experienced any event by time~$h$ (corresponding to the condition that $T\ge h$). Fine and Gray added the extra unnatural condition (roughly, a subject could also be at risk of event $k$ happening even if they already experienced a different event earlier) primarily for modeling convenience; this choice makes interpreting the subhazard difficult but turns out to make interpreting the impact of features on CIFs more straightforward, as we explain shortly.

With the above definition of a subhazard, the Fine and Gray model applies Cox's proportional hazards assumption to the subhazard as to get a \emph{proportional subhazards assumption}: %
\begin{equation}
\lambda _k (h|x) = \lambda_{k,0}(h) \exp\bigl\{\phi_k (x; \theta_k)\bigr\}\qquad\text{for all }h\ge0, x\in\mathcal{X},
\label{eq:prop-hazards}
\end{equation}
where $\lambda_{k,0}(h)$ is the baseline subhazard function for event $k$; and $\phi_k(\cdot;\theta_k)$ is a transformation on the feature vector $x$ with learnable parameter $\theta_k$.
In their original paper, Fine and Gray assumed $x$ and $\theta_k$ to be Euclidean vectors of the same length with $\phi_k(x;\theta_k)=x^\top \theta_k$, whereas in this paper we more generally let $\phi_k$ be a user-specified neural net (that maps an input feature vector from the input space $\mathcal{X}$ to a real-valued ``risk score'').

\paragraph{Implications of the proportional subhazards assumption}
Under the proportional subhazards assumption, \citet{fine1999proportional} showed that if we apply the transformation $g(u)\triangleq\ln \bigl\{-\ln (1-u)\bigr\}$ to the CIF, we get
\[
g\big(F_k(h|x)\big) = \mu_0(h) + \phi_k(x; \theta_k),
\]
where $\mu_0(h) = \ln \bigl\{\int_0^h \lambda_{k,0}(\eta)d\eta\bigr\}$. 
Since $g$ is a monotonically increasing function, the main takeaway here is that as $\phi_k(x; \theta_k)$ increases, so does the CIF value $F_k(h|x)$. This interpretation is a key reason why the Fine and Gray model is popularly used, despite its subhazard being uninterpretable. If we were to remove the unnatural condition from equation~\eqref{eq:subhazard}, we would no longer have a ``clean'' relationship between the subhazard for a specific event~$k$ and event $k$'s CIF because event $k$'s CIF will depend on subhazards for all events, not just that of event $k$ (for more details, see, for instance, Section 12.3.2 of \citet{collett2023modelling}).

To further emphasize the ease of interpreting a Fine and Gray model, consider any two feature vectors $x,x'\in\mathcal{X}$. We can compare the log subhazards ratio between $x$ and $x'$:
\begin{align}
\ln \frac{\lambda_k(h|x')}{\lambda_k(h|x)} 
&= \ln \lambda_k(h|x') - \ln \lambda_k(h|x) \nonumber \\
&= \ln \lambda_{k,0}(h) + \phi_k(x'; \theta_k) - \ln \lambda_{k,0}(h) - \phi_k(x; \theta_k)\qquad\text{(plugging in equation~\eqref{eq:prop-hazards})} \nonumber \\
&= \phi_k(x'; \theta_k) - \phi_k(x; \theta_k).
\label{eq:loghazard-convert}
\end{align}
The right-hand side has a straightforward interpretation: how much more is the risk of event~$k$ for $x'$ compared to $x$. A positive log ratio means $x'$ has a higher risk than $x$ for event $k$; a negative log ratio means the opposite. The ease of interpreting the log subhazards ratio and the monotonic relationship between an event's subhazard and the same event's CIF are the key reasons our proposed approach builds on the Fine and Gray model.

\paragraph{Model training} The parameter $\theta_k$ could be learned by maximizing a partial likelihood function with inverse probability of censoring weighting (IPCW) adjustment due to competing risks, while the baseline subhazard $\lambda_{k,0}(h)$ can be estimated using Breslow's method, also with IPCW adjustment. As both of these steps are standard and explaining them is unnecessary for understanding the high-level ideas of our proposed method, we defer presenting how the Fine and Gray model training works to Appendix~\ref{app:fine-gray-model}.
In summary, after model training, per event $k$, we have both an estimate $\widehat\theta_k$ of $\theta_k$ (neural net parameters) and an estimate $\widehat\lambda_{k,0}$ of the baseline subhazard $\lambda_{k,0}$.

\paragraph{Prediction} Using the learned model parameters, for test feature vector $x\in\mathcal{X}$, we can calculate the predicted CIF for event $k\in[m]$ as
\begin{align} \label{eq:cif-estimate}
\widehat F_k(h|x) = 1 - \exp\bigl\{-\exp\{ \phi_k(x;  \widehat \theta_k)\} \widehat \Lambda_{k,0}(h)\bigr\}\qquad\text{for }h\ge0,
\end{align}
where
$\widehat \Lambda_{k,0}(h) = \int_0^h \widehat \lambda_{k,0}(\eta) d\eta$ is called the \emph{baseline cumulative subhazard}.

\paragraph{Other competing risks models}
Recently, there are other competing risks models \citep{lee2018deephit, nagpal2021deep, jeanselme23a, alberge2025survival} proposed. However, to the best of our knowledge, they do not provide a straightforward interpretation of how the risk of an event differs between two feature vectors (of a form like in equation~\eqref{eq:loghazard-convert}), which we crucially build upon to check for whether adding new features benefits prediction. Moreover, these other flexible models allow for complex interactions between the critical events so that the CIF of one event could depend on information from all events.

Separately, when considering the specific medical problem we focus on (prognostication of comatose post-cardiac arrest patients) albeit using different features, existing work by \citet{shen23neuro} (who used EEG data, whereas we focus on hemodynamic data) showed that the Fine and Gray model is a strong baseline, with model performance on par with or exceeding more recently proposed deep learning methods (namely, Dynamic-DeepHit \citep{lee2019dynamic} and DDRSA \citep{venkata2022intervene}).

\section{Proposed Method: Stepwise Fine and Gray}\label{sec:method}

We now describe our proposed model, Stepwise Fine and Gray, that can evaluate the prognostic impact of hemodynamic signals on neurological outcomes in comatose post-cardiac arrest patients. Our approach builds upon the Fine and Gray model described in Section~\ref{sec:fg-model} and employs neural networks to flexibly parameterize the effects of features. In what follows, we describe a two-phase learning procedure and a thresholding strategy to assess the incremental contribution of time-varying features. Accommodating more than two phases in which new features are collected is straightforward; as this generalization is not essential to understanding our proposed approach nor our experiments, we discuss it in Section~\ref{sec:discussion}. For our specific application, we use $m=3$ competing events: awakening from coma, death following WLST, and death despite maximal support.

\paragraph{The dynamic prediction task}
Following the notation in Section~\ref{sec:background} but omitting subject index $i$ for simplicity, we now add time index $t$ to capture a patient's varying features over time after the cardiac arrest. Specifically, let $X_{t}$ denote the feature vector for subject $i$ at prediction time $t$ ($X_{t}$ can consist of both static and time-varying features). In our context, the prediction time~$t$ represents the time after a patient's cardiac arrest onset (i.e., time 0 is when the patient experienced cardiac arrest). As $t$ increases (in which we could collect more data from the patient), our goal is to predict the time until the earliest competing event as well as which event it is, starting from time $t$ for the patient.

For ease of exposition, in this section, it suffices to think of $t$ as fixed. For example, if $t$ is set to be 12 hours, then it means that for every subject, we use their first 12 hours of collected data (summarized into a fixed-length feature vector) to predict the time until the earliest competing event as well as which event it is, measured starting from hour 12. Formally, we predict the CIF of each event $k\in[m]$, where time horizon $h>0$ is measured starting from hour 12:
\[
F_k(h|x_t) \triangleq \mathbb{P}(T\le h, D=k \mid X_t = x_t).
\]
This is a slightly modified version of equation \eqref{eq:cif}. The only modification is that we explicitly indicate that the feature vector is based on data collected up to time $t$ for the subject.

In our experiments later, we vary $t$. A single model is trained that works for all $t$, rather than re-training a separate model for different values of $t$. The details of handling varying values of $t$ are not essential to understanding the key ideas of our proposed method and are deferred to Appendix~\ref{app:model-t}.

\paragraph{Two-phase learning}
We decompose $X_{t}$ into two components $X_{t} = \bigl(X_{t}^{(1)},\, X_{t}^{(2)}\bigr)$. Here, $X_{t}^{(1)}$ includes static features (collected early on) as well as the value of $t$ itself (i.e., time elapsed since cardiac arrest). Including $t$ as part of $X_{t}^{(1)}$ is a basic way to capture time-dependent variation, assuming no other time-varying features were available (e.g., the perceived probability of awakening for a patient who remains in a coma 8 hours after cardiac arrest might be different from a patient who remains in a coma 24 hours after cardiac arrest, even if they have the exact same static features initially collected and assuming that we know nothing else about the patients). Next, $X_{t}^{(2)}$ consists of time-varying hemodynamic features (e.g., blood pressure trajectories, vasopressor usage) summarized at time~$t$.

Our goal is to update risk predictions for different events by first modeling the baseline risk with $X_{t}^{(1)}$ (Phase 1) and then incorporating the additional information from $X_{t}^{(2)}$ (Phase~2). We conduct the two-phase learning for each competing event of interest. In what follows, we focus on the $k$-th event; the same procedure applies to other events as~well.

Our method sequentially learns two Fine and Gray models, the latter building on the first (hence why we call our method \emph{Stepwise Fine and Gray}).

\paragraph{Phase 1 (static features):}
In the first phase, we train a Fine and Gray model using only features $X_{t}^{(1)}$. Specifically, the subhazard function for event $k$ is
\begin{align}
  \lambda_{k,t}^{(1)}(h \mid X_{t}^{(1)}) = \lambda_{k,0}^{(1)}(h)\exp\Bigl\{ \phi_{k,t}^{(1)} \bigl(X_{t}^{(1)};\, \theta_k^{(1)}\bigr)\Bigr\},
\end{align}
where $\lambda_{k,0}^{(1)}(h)$ is the baseline subhazard. %
 After learning the model, we obtain the estimated log partial subhazard
$\widehat{f}_{k,t}^{(1)}(X_{t}) = \phi_{k,t}^{(1)}\bigl(X_{t}^{(1)};\, \widehat{\theta}_k^{(1)}\bigr)$, i.e., $\widehat{\theta}_k^{(1)}$ contains the learned neural net parameters. We can think of $\widehat{f}_{k,t}^{(1)}(X_{t})$ as the risk score for $X_{t}$ only using Phase 1 features.

\paragraph{Phase 2 (time-varying features):}
In the second phase, we incorporate the time-varying features $X_{t}^{(2)}$ along with $X_{t}^{(1)}$ to refine the risk prediction. Specifically, we define a second risk function
$\phi_{k,t}^{(2)}\bigl(X_{t}^{(1)},X_{t}^{(2)};\, \theta_k^{(2)}\bigr)$
with parameters $\theta_k^{(2)}$. We still use a Fine and Gray model except where the overall subhazard function is now taken to be
\begin{align}
  \lambda_{k,t}^{(2)}(h \mid X_{t}^{(1)},X_{t}^{(2)}) = \lambda_{k,0}^{(2)}(h)\exp\Bigl\{ \underbrace{\widehat{f}_{k,t}^{(1)}(X_{t})}_{\text{treated as fixed}} +~ \phi^{(2)}\bigl(X_{t}^{(1)}, X_{t}^{(2)};\, \theta_k^{(2)}\bigr)\Bigr\},
\end{align}
where $\lambda_{k,0}^{(2)}(h)$ is a different baseline subhazard estimated in Phase 2. Note that the risk prediction from Phase 1 is treated as fixed (so $\widehat\theta_k^{(1)}$ is frozen in Phase 2 model training). The combined risk prediction is then given by
$\widehat{f}_{k,t}^{(1)}(X_{t}) + \widehat{f}_{k,t}^{(2)}(X_{t})$ with $\widehat{f}_{k,t}^{(2)}(X_{t}) = \phi_{k,t}^{(2)}\bigl(X_{t}^{(1)}, X_{t}^{(2)};\, \widehat\theta_k^{(2)}\bigr).$
By using the proportional subhazards assumption, the log subhazards ratio (similar to equation~\eqref{eq:loghazard-convert}) that compares the Phase~2 model to the Phase~1 model is
\begin{align}
I_k(h|X_t) &\triangleq
\ln\frac{\widehat{\lambda}_k^{(2)}(h \mid X_{t}^{(1)}, X_{t}^{(2)})}{\widehat{\lambda}_k^{(1)}(h \mid X_{t}^{(1)})}
\nonumber \\ %
&= \big[\ln \widehat\lambda_{k,0}^{(2)}(h) + \widehat{f}_{k,t}^{(1)}(X_t) + \widehat{f}_{k,t}^{(2)}(X_t)\big] - \big[\ln \widehat\lambda_{k,0}^{(1)}(h) + \widehat{f}_{k,t}^{(1)}(X_t)\big] \nonumber \\
&= \widehat{f}_{k,t}^{(2)}(X_{t}) + \ln \widehat\lambda_{k,0}^{(2)}(h) - \ln \widehat\lambda_{k,0}^{(1)}(h).
\label{eq:log-subhaz-ratio-increment}
\end{align}
We refer to the log ratio $I_k(h|X_t)$ as the \emph{incremental contribution} of the Phase 2 features (in our application, the time-varying hemodynamic features) over the Phase 1 features (static features). %
Instead of comparing the risks of two feature vectors that correspond to two subjects (as in equation~\eqref{eq:loghazard-convert}), $I_k(h|X_t)$ compares the risks of the same subject with access to two different sets of features $X_t^{(1)}$ vs $X_t = (X_t^{(1)},X_t^{(2)})$. The monotonic relationship between an event's subhazard and CIF (Section~\ref{sec:fg-model}) ensures that positive log subhazard ratios correspond to increased event probabilities. Thus, when $I_k(h|X_t)$ is positive, the inclusion of Phase 2 features is associated with increased predicted probabilities of the event occurring for the subject corresponding to feature vector $X_t$, while a negative value suggests the opposite.

For instance, when modeling a favorable event (e.g., a coma patient waking up), a positive $I_k(h|X_t)$ means that the patient's Phase 2 features are associated with an improved chance of awakening. Conversely, when modeling an adverse event (e.g., death), a positive $I_k(h|X_t)$ suggests that the patient's Phase 2 features are linked to a higher risk of death.

\paragraph{Thresholding the incremental contribution}
To determine when Phase 2 features $X_{t}^{(2)}$ significantly improve prediction, we introduce a thresholding procedure on the incremental contribution $I_k(h|X_t)$. We learn a threshold across $\delta_k(h)$, which is not patient-dependent, by selecting the value that yields the best performance on a held-out validation set. In particular, if $|I_k(h|X_t)|>\delta_k(h)$, then we use the Phase 2 model to predict. Otherwise, we fall back to the Phase 1 model. Importantly, this form of variable selection (when to use only Phase 1 features vs features from both phases) depends on $X_t$ and thus is patient-specific.

\paragraph{Model training}
As we have described in Section~\ref{sec:fg-model}, parameters $\theta_k^{(1)}$ and $\theta_k^{(2)}$ can be separately estimated by maximizing the IPCW-adjusted partial likelihood as defined in equation~\eqref{eq:ipcw-ll}. We choose to use neural network architectures (e.g., a few linear layers with some nonlinear activation functions) to parameterize $\phi_{k,t}^{(1)}$ and $\phi_{k,t}^{(2)}$ to capture non-linear relationships among features. The baseline hazards $\lambda_{k,0}^{(1)}$ and $\lambda_{k,0}^{(2)}$ are then estimated using an IPCW-adjusted Breslow estimator. The cumulative baseline subhazard functions $\Lambda_{k,0}^{(1)}(h)$ and $\Lambda_{k,0}^{(2)}(h)$ are then obtained by integrating the estimated baseline hazards. We tune the threshold function $\delta_k(h)$ on a validation set.

\paragraph{Prediction} 
Once we have learned the parameters $\widehat\theta_k^{(1)}$, $\widehat\theta_k^{(2)}$ and the baseline subhazard functions, we can calculate $\widehat f_{k,t}^{(1)} = \phi_{k,t}^{(1)} \bigl(X_{t}^{(1)};\, \widehat\theta_k^{(1)}\bigr)$ and $\widehat f_{k,t}^{(2)} = \phi_{k,t}^{(2)} \bigl(X_{t}^{(1)}, X_{t}^{(2)};\, \widehat\theta_k^{(2)}\bigr)$, and the Phase 1 and Phase 2 CIF estimates $\widehat F_{k,t}^{(1)}(h \mid X_t)$ and $\widehat F_{k,t}^{(2)}(h \mid X_t)$, respectively, for the $k$-th event at time $t$ for the two phases respectively using equation~\eqref{eq:cif-estimate}: 
\begin{align*}
\text{CIF from Phase 1: }& \widehat F_{k,t}^{(1)}(h \mid X_t) = 1 - \exp\bigl\{-\exp\bigl\{ \widehat f_{k,t}^{(1)} \bigr\} \widehat \Lambda_{k,0}^{(1)}(h)\bigr\} \\ 
\text{CIF from Phase 2: }& \widehat F_{k,t}^{(2)}(h \mid X_t) = 1 - \exp\bigl\{-\exp\bigl\{ \widehat f_{k,t}^{(1)} +  \widehat f_{k,t}^{(2)} \bigr\} \widehat \Lambda_{k,0}^{(2)}(h)\bigr\} 
\end{align*}
Letting $\widehat{\delta}_k(h)$ denote the learned thresholding function, the final predicted CIF for event $k$ at time $t$ is
\begin{equation}
  \widehat F_{k,t}(h \mid X_t) = 
  \begin{cases}
    \widehat F_{k,t}^{(1)}(h \mid X_t) & \text{if } |I_k(h|X_t)|\le\widehat{\delta}_k(h), \\
    \widehat F_{k,t}^{(2)}(h \mid X_t) & \text{if } |I_k(h|X_t)| >\widehat{\delta}_k(h),
  \end{cases}
  \label{eq:threshold}
\end{equation}
where $I_k(h|X_t)$ is computed using the right-hand side of equation~\eqref{eq:log-subhaz-ratio-increment}.
Although the thresholding rule yields a binary choice between Phase 1 and Phase 2 at any fixed prediction time $t$, the timing of that choice is dynamic and patient-specific. For each patient, event $k$, and horizon $h$, we re-evaluate the incremental contribution as new data arrive and $t$ advances.

Overall, our two-phase competing risks method enables us to see for whom (i.e., for which $X_t$) and also when (since $X_t$ depends on $t$) it suffices to only consider Phase 1 (static) features versus also considering Phase 2 (time-varying) features (since the thresholding in equation~\eqref{eq:threshold} depends on $I_k(h|X_t)$), providing insight into the incremental value of continuous hemodynamic monitoring in the context of prognostication.

\section{Cohort}\label{sec:cohort}
We now describe the cohort of comatose post-cardiac arrest patients and the process for data extraction.

\paragraph{Cohort selection} 

We retrospectively identified comatose post-cardiac arrest patients admitted to the intensive care unit (ICU) at a large medical center in the Northeastern US between 2010 and 2022. Inclusion criteria required that patients be adults who experienced a cardiac arrest, were successfully resuscitated, and remained comatose upon ICU admission. Patients withdrawn from life-sustaining therapy for non-neurological reasons (e.g., they never wished to undergo ICU care) were excluded. Additionally, since our goal is to assess the incremental clinical value of time-varying features beyond the static baseline features collected within 6 hours post-arrest, we excluded patients whose first event occurred during this period. After applying these criteria, the final analytic cohort consisted of 2,278 patients (see summary statistics in Appendix~\ref{app:data}). As already stated at the start of Section~\ref{sec:method}, we consider three competing events: awakening from coma, death following WLST, and death despite maximal support. For patients experiencing multiple events (e.g., awakening followed by WLST), we use the earliest event. As a robustness check, we repeated our analysis excluding patients admitted in 2020 and 2021 (corresponding to the COVID-19 pandemic); our findings remain largely the same (see Appendix~\ref{app:exclude-covid} for details).

\paragraph{Data extraction} 

A set of clinical variables was extracted from the electronic health record and a prospective registry maintained at this institution. Demographic information (e.g., age, gender) and cardiac arrest characteristics (e.g., location of arrest, transfer status, arrest rhythm) were obtained. Early neurological exam results, including the motor component of the Full Outline of Unresponsiveness (FOUR) score, as well as pupillary and corneal reflex responses, were recorded within the first 6 hours post-arrest. 
Longitudinal hemodynamic data were also collected, including MAP measurements and vasopressor dosages. 
In addition, the time of first awakening (i.e., when a patient was first able to follow commands) and the time of death (whether following WLST or despite maximal support) were recorded, allowing us to calculate the time-to-event for each of the three competing outcomes. 

Time-varying data were aggregated into 1-hour snapshots up to the occurrence of the first event. Missing values in the mean blood pressure (BP) series were imputed using the last observation carried forward method, and summary statistics (mean, minimum, and maximum) were computed for each 1-hour window. For vasopressor dosages, both the hourly mean and cumulative dosage since arrest were calculated. Time elapsed since the cardiac arrest was included as an additional feature.

Static features collected within the first 6 hours post-arrest and time elapsed since arrest were used in Phase 1 of our analysis, while Phase 2 incorporated both static and dynamic features from the hourly snapshots. %
A full list of features and their processing details is provided in Appendix~\ref{app:data}.

\section{Experiment Results} 
We trained our stepwise Fine and Gray model on the cohort described in Section~\ref{sec:cohort}. The models are trained using data up to 10 days post‐arrest (for computational efficiency, we did not consider more than 10 days). A stratified 64/16/20 train/validation/test split based on patients was used (stratified by the motor component of the FOUR score), and experiments were repeated 5 times to ensure robustness. For threshold tuning on the incremental contribution from Phase 2, for each horizon value $h\in\{24,48,72,120,240\}$ (hours), we select the threshold that yields the best validation performance. The search grid consists of 100 uniformly sampled values ranging from 0 up to the maximum absolute value of the incremental contribution. See Appendix~\ref{app:experiment} for additional experimental details. As explained in Appendix~\ref{app:model-t}, during learning, we use instances from training patients by sampling hourly snapshots with a step size of 5-hour and train the models separately for each of the three competing events we mentioned: awakening from the coma for the first time, death despite maximal support (i.e., not from WLST), and death from WLST. 

In what follows, we first describe the evaluation approach and model performance. We also look at individual-level and subgroup analysis to further understand the clinical insights from our stepwise Fine and Gray model.

\subsection{Evaluation Approach} 

To evaluate the model’s discriminative performance, we used the concordance index adapted for competing risks (CR c-index) following \cite{wolbers2014concordance} and \cite{ishwaran2014random}.
The CR c-index for the $k$-th event is defined as
\begin{align}
    c_k = \mathbb{P}\bigl\{ \widehat f_k(X_i | \widehat{\theta_k}) >  \widehat f_k(X_j | \widehat{\theta_k}) \mid D_i=k \text{ and } (Y_i < Y_j \text{ or }  D_j \neq k)  \bigr\}
\end{align}
which accounts for the presence of competing events. It falls in the range of $[0,1]$, with 1 indicating a perfect separation on patients' risk. For reference, we also report the classic Harrell c-index \citep{harrell1982evaluating} in Appendix~\ref{app:experiment}, though it is known to be biased when considering competing risks by treating them as censored events~\citep{wolbers2014concordance}.

\subsection{Model Performance} 

We report the discriminative performance of our stepwise Fine and Gray model, as measured by the competing risks (CR) c-index, for each event at various prediction times $t$ (hours since cardiac arrest, as described in Section \ref{sec:method}) in Table~\ref{tab:cindex-2nnfg}. At each time $t$, only patients who remain under observation (i.e., those still in a coma $t$ hours after cardiac arrest) are included; patients who have experienced an event (awakening or death) before that time are excluded from subsequent evaluations (see Figure~\ref{fig:num_patients_at_t} for details). More details of the threshold tuning procedure can be found in Appendix~\ref{app:experiment}.

\begin{table}[htbp]
  \centering
  \caption{CR c-index of the Two-Phase Neural Network Fine and Gray Model. ``Phase~1'' refers to predictions based solely on static features; ``Phase~2'' incorporates time-varying hemodynamic features; ``Tuned~Threshold'' applies a threshold to the Phase 2 contribution to determine its clinical significance. Values are reported as mean $\pm$ standard deviation over 5 experimental repeats.} \label{tab:cindex-2nnfg} 
  \vspace{-0.5em}
  \begin{adjustbox}{max width = 0.85\textwidth}
  \begin{tabular}{ccccc}
    \toprule
    Event & Prediction Time $t$ (hr) & Phase 1 & Phase 2 & Tuned Threshold \\
    \midrule
      \multirow{4}{*}{Awakening} & 6 & \bftab{0.9253$\pm$0.0136} & 0.8298$\pm$0.0366 & 0.9238$\pm$0.0138 \\ 
      ~ & 12 & \bftab{0.9211$\pm$0.0060} & 0.8260$\pm$0.0169 & 0.9205$\pm$0.0044 \\ 
      ~ & 24 & \bftab{0.9132$\pm$0.0091} & 0.7864$\pm$0.0366 & 0.9039$\pm$0.0145 \\ 
      ~ & 48 & \bftab{0.8920$\pm$0.0126} & 0.7452$\pm$0.0234 & 0.8837$\pm$0.0176 \\ 
    \midrule
      & 6 & 0.7773$\pm$0.0225 & 0.8146$\pm$0.0139 & \bftab{0.8415$\pm$0.0191} \\ 
      non-WLST & 12 & 0.7816$\pm$0.0164 & 0.8326$\pm$0.0302 & \bftab{0.8470$\pm$0.0230} \\ 
      Death & 24 & 0.7859$\pm$0.0283 & 0.7927$\pm$0.0238 & \bftab{0.8282$\pm$0.0260} \\ 
      ~ & 48 & 0.8094$\pm$0.0427 & 0.7542$\pm$0.0504 & \bftab{0.8258$\pm$0.0382} \\ 
    \midrule
      \multirow{4}{*}{WLST} & 6 & 0.8026$\pm$0.0277 & 0.7313$\pm$0.0549 & \bftab{0.8076$\pm$0.0271} \\ 
      ~ & 12 & 0.8292$\pm$0.0272 & 0.7388$\pm$0.0529 & \bftab{0.8297$\pm$0.0232} \\ 
      ~ & 24 & \bftab{0.8522$\pm$0.0275} & 0.7368$\pm$0.0495 & 0.8488$\pm$0.0236 \\ 
      ~ & 48 & \bftab{0.8562$\pm$0.0261} & 0.7476$\pm$0.0449 & 0.8545$\pm$0.0262 \\ 
      \bottomrule
  \end{tabular} 
  \end{adjustbox}
  
\end{table}

For the event of awakening, predictions based solely on static features yield high performance (Phase~1 column in Table~\ref{tab:cindex-2nnfg}). The addition of time-varying hemodynamic features, when incorporated without thresholding, tends to lower the c-index (Phase~2 column in Table~\ref{tab:cindex-2nnfg}); however, when a threshold is applied to the Phase~2 contribution, performance is nearly as good as that of using static features alone (Tuned~Threshold column in Table~\ref{tab:cindex-2nnfg}). Thus, hemodynamic features should be carefully integrated to avoid decreased discrimination performance.

For the event of death not from WLST, the inclusion of time-varying hemodynamic data significantly improves prognostication, suggesting that blood pressure (BP) information is important in predicting fatal outcomes. Furthermore, tuning the threshold for the Phase~2 contribution further enhances model discrimination. In a later subgroup analysis (Section~\ref{sec:subgroup-analysis}), we provide more insights on what clinical prognostic value it brings from including BP and vasopressor dosage information. 

For the WLST event, static features strongly predict physicians’ decisions, and incorporating time-varying features without thresholding appears to diminish performance. 
This suggests that the key baseline factors influencing WLST decisions, such as patient demographics, time elapsed since arrest, and the initial neurological examinations, are already captured by Phase 1 features.  These findings are consistent with the cognitive models reported in \cite{steinberg2022physicians}, where physicians emphasized the importance of these foundational factors in their decision-making process.

Naively using the Phase~2 model everywhere can lower discrimination performance for two reasons. Technically, the Phase~2 model is learned with the Phase~1 model held fixed, which restricts flexibility; combined with noisy hemodynamic inputs, this can reduce performance. Clinically, hemodynamics are volatile and intervention-driven, so brief changes in blood pressure and vasopressor dosage may not indicate whether a patient’s brain function is improving or declining in the long term. Our thresholding therefore uses the Phase~2 model only when $|I_k(h\mid X_t)|>\hat\delta_k(h)$, preserving the stronger Phase~1 signal when the additional hemodynamic monitoring does not significantly help. We show the distribution of the incremental contribution from Phase 2 and the learned thresholds for different events from one repeat of the experiment in Appendix~\ref{app:experiment}.

We also observe that the performance trends vary with the prediction time. For awakening and non-WLST death, the CR c-index tends to decrease as prediction time increases, suggesting that patients remaining in a coma for longer periods represent more challenging cases for neuroprognostication. 
In contrast, performance for WLST increases over time, which is consistent with the notion that static features are strong predictors of physicians’ decisions regarding WLST.

We also compare our model against baselines DeepSurv \citep{faraggi1995neural, katzman2018deepsurv}, DeepHit \citep{lee2018deephit}, and Deep Survival Machines \citep{nagpal2021deep}, reporting achieved test set CR c-index scores in Table~\ref{tab:cindex-cr-sota} in Appendix~\ref{app:experiment}. Our stepwise Fine and Gray model achieves similar or higher performance while providing additional interpretation on the incremental contribution from Phase 2 features.

\subsection{Individual-level Visualization}
\begin{figure}[ht]
  \centering
  \includegraphics[width=1\linewidth]{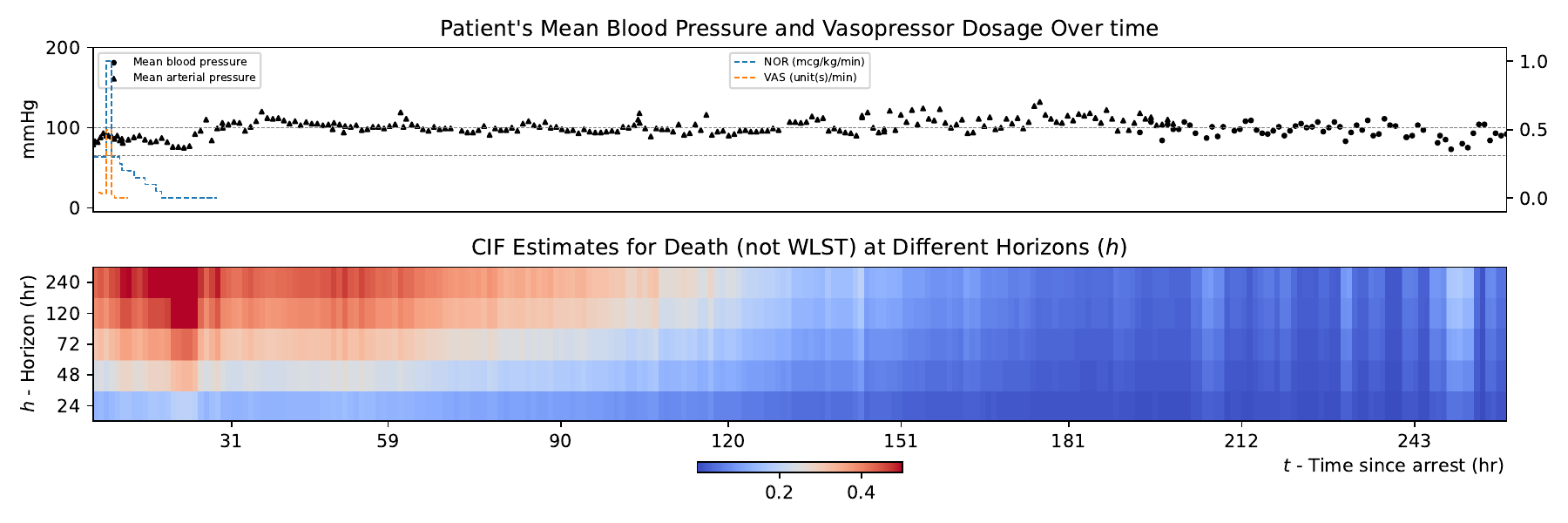}
  \vspace{0.em}
  \includegraphics[width=1\linewidth]{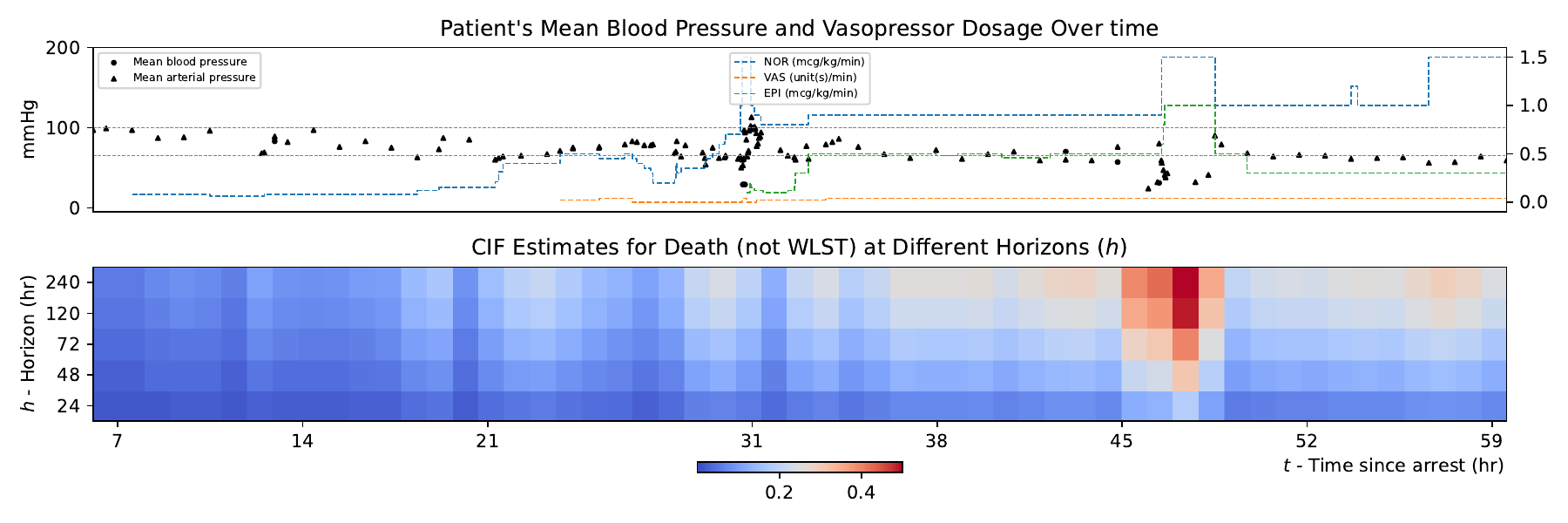}
  \vspace{-2em}
  \caption{\textbf{Top Panel:} A patient with an initially poor neurological exam who ultimately awakened; the gradual decrease in predicted risk reflects benign hemodynamic trends. \textbf{Bottom Panel:} A patient with a moderate early neurological exam who died despite maximal support; notable drops in predicted risk coincide with MAP falling below 65 mmHg. A 65-100 mmHg interval for mean BP marked with two dashed gray horizontal lines represents a recommended range for mean BP.}
  \vspace{-1em}
  \label{fig:patient-visualizations}
\end{figure}

We develop individual visualizations using patient trajectory and predicted CIF for the event of death (not from WLST), as the time-varying features tend to give the greatest value on the event of non-WLST death compared to the other two events.
Figure~\ref{fig:patient-visualizations} illustrates two representative patient trajectories for the event of death (excluding WLST). In the top panel, a patient with a poor initial neurological exam (no response to pain) eventually awakened after approximately 10 days. The predicted risk of death gradually decreased over time, reflecting stable mean arterial pressure readings. In contrast, the bottom panel shows a patient with a relatively favorable early exam (flexion response to pain) who died after about 3 days despite maximal support; here, marked drops in the predicted risk were observed when blood pressure fell below 65 mmHg, particularly during later periods.

\subsection{Subgroup Analysis}\label{sec:subgroup-analysis}
We performed subgroup analyses based on the motor component of the FOUR score, categorizing patients into three groups: (1) Group 1: No response or extension to pain, or myoclonus – representing severe brain dysfunction; (2) Group 2: Flexion to pain – intermediate impairment; (3) Group 3: Localizing to pain – least severe dysfunction.

\begin{figure}[ht]
  \centering 
  \includegraphics[width=1.\linewidth]{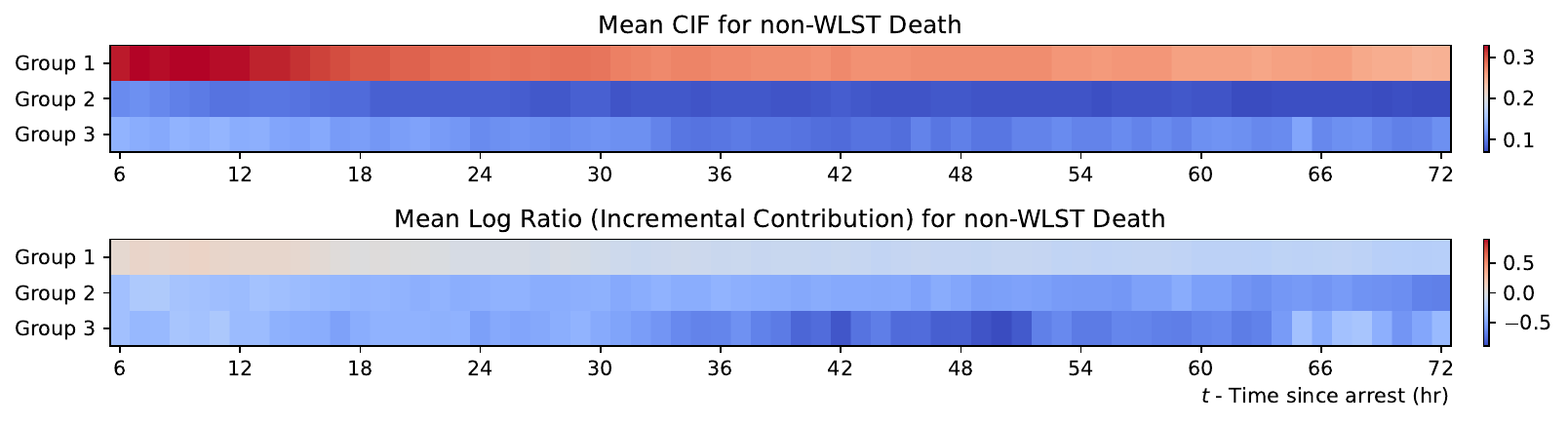}
  \vspace{-1em}
  \caption{
  Average CIF at $h=240$ hours and Phase 2 Log Subhazard Contributions by Neurological Exam Subgroup. The mean predicted CIFs at a $h=10$ horizon (i.e., probability of death within 10 days of arrest) and the corresponding log subhazards ratio (i.e., the incremental contribution of hemodynamic features $I(240|X_t)$ using equation~\eqref{eq:log-subhaz-ratio-increment}) are shown for three subgroups defined by the motor component of the FOUR score. The results indicate that patients with the most severe neurological dysfunction (Group 1) exhibit the smallest incremental prognostic value from dynamic features, whereas patients with intermediate (Group 2) and less severe impairment (Group 3) demonstrate larger positive contributions, suggesting that continuous cardiovascular monitoring is more informative for those with relatively preserved neurological function.}
  \label{fig:subgroup-cif-s2} 
\end{figure}

For each subgroup, we computed the average CIF at $h=240$ (10 days after the prediction time $t$), which represents the probability of death despite maximal support, and the mean incremental contribution (defined in equation~\eqref{eq:log-subhaz-ratio-increment}) from Phase~2, which reflects the relative effect of time-varying features compared to the static baseline prediction from Phase 1.  It is important to emphasize that the Phase 2 incremental contribution reflects changes from the Phase 1 risk prediction; for example, in Group 1 patients, whose Phase 1 risk prediction are already very high due to severe neurological impairment, even a modest contribution from time-varying features may result in only a small or even positive incremental contribution to the risk of death.
Note that only patients who had not yet experienced an event and remained in a coma at each time point were included in these averages. 
Figure~\ref{fig:subgroup-cif-s2} displays these subgroup comparisons, with predictions evaluated up to 72 hours post-arrest.

Our subgroup analysis reveals a clear trend in the mean CIF for death (not from WLST) across different neurological impairment groups. Specifically, Group 1, the most severely impaired, exhibits the highest mean CIF, indicating the poorest prognosis, while Groups 2 and 3 demonstrate lower CIF values, consistent with their comparatively better outcomes.

In addition, incorporating blood pressure and vasopressor dosage information over time generally results in lower risks of death (i.e., negative log subhazard ratios) when averaged across patients still under observation, suggesting that the dynamic data tend to provide favorable prognostic signals. However, for Group 1 patients, there is an early period (approximately the first 18 hours since the arrest) during which the incremental contribution becomes positive, suggesting an increased risk of death associated with the dynamic features during that window. 
This finding aligns with clinical expectations: in patients with severe brain dysfunction, the potential for recovery is limited, so additional cardiovascular information may offer less prognostic value. 
In contrast, patients in Groups 2 and 3, who have lower baseline risks from static features, exhibit a more pronounced negative incremental contribution, indicating a stronger beneficial effect from continuous hemodynamic monitoring. 

Moreover, the trend toward increasingly positive incremental effects over time may reflect that patients who persist in a coma at later time points, having avoided early adverse events, are more likely to have a better chance of recovery.

\section{Discussion} 
\label{sec:discussion}

Our stepwise Fine and Gray model integrates static baseline features and time-varying hemodynamic features to enhance prognostication in comatose post‐cardiac arrest patients. Our results demonstrate that incorporating dynamic data significantly improves risk prediction, particularly for death despite maximal support. While early neurological assessments remain robust predictors for awakening and decisions regarding the withdrawal of life-sustaining therapy, the additional information from continuous hemodynamic monitoring refines risk estimates over time. By decomposing the overall risk into static and dynamic components via the additive log-subhazard property, our method provides interpretable risk estimates. Notably, subgroup analyses based on the motor component of the FOUR score reveal that the incremental prognostic value of time-varying features differs among patients; those with moderate-to-mild neurological impairment benefit more from continuous monitoring compared to patients with severe dysfunction. This observation underscores the potential for tailoring monitoring strategies to individual patient profiles.

\paragraph{Extension to more phases}
We motivated our stepwise Fine and Gray model using two phases with a real medical application where, in practice, Phase~1 features are collected prior to Phase~2 features. There could be situations where we start collecting yet another set of features (such as EEG data) that could be thought of as Phase 3 features. We could repeat the same strategy as our two-phase approach of Section~\ref{sec:method}, treating Phase~1 and Phase~2 features collectively as what we previously referred to as Phase~1 features, while treating the Phase~3 features as what we previously called the Phase~2 features. In this manner, our approach can easily extend to an arbitrarily large number of phases that correspond to newly collected features over time.

As another modeling choice, we had motivated using hemodynamic data as Phase~2 features such as mean blood pressure and vasopressor dosages. We could separate these out into, for instance, Phase~2 (mean blood pressure) and Phase~3 (vasopressor dosages) features; even though we might start collecting these in tandem in the ICU for different patients, we may want to tease apart the contribution of these two separately (where it is possible to see how much Phase~2 contributes over Phase~1, how much Phase~3 contributes over Phase~1, and also how much Phases 2~and~3 together contribute over Phase~1 in an ablation-style analysis). Making phases too fine-grained could make it so that each new phase does not contribute enough new information per subject. Making a phase too coarse could make it so that we are grouping together too many features into the phase, making it harder to interpret what is special about a phase even if it significantly helps prediction. We have thematically grouped together all hemodynamic features into a single Phase~2 with the idea that clinicians would find this sort of grouping still intuitive without making the resulting model overly complicated.

\paragraph{Limitations}
We discuss some limitations of our work.
First, our study focuses on hemodynamic and early neurological exam data. Future studies could integrate additional modalities (e.g., EEG, serial neurological exams, and brain imaging) to provide a more comprehensive assessment.
Second, for simplicity, we did not incorporate dedicated time-series encoders (e.g., RNNs, transformers) and instead focused on an interpretable variable selection approach to derive medical insights; although integrating such encoders is straightforward, they may capture more intricate temporal dynamics. 
Third, while MAP is continuously monitored in clinical practice, our dataset captures near-hourly recordings, and our imputation strategy relies on the last observation carried forward; more granular MAP data could potentially enhance model performance. 
Fourth, our method builds on the Fine and Gray model, which has some limitations by itself. For example, the CIFs generated by the model do not necessarily sum to one across events at some maximal horizon. Although normalization approaches have been proposed \citep{jeanselme23a}, they risk disrupting the proportional properties and the fundamental mathematical relationships between an event's subhazard and its CIF that underpin the original model, so we chose not to adopt them. Additionally, we are using it for a dynamic prediction setup, which is not what Fine and Gray was initially designed for (see Appendix~\ref{app:model-t} for more explanations on this).
Finally, it is important to note that our clinical insights are derived from observational data collected at a single medical center. As such, further medical studies and external validation are required to verify and generalize these findings across diverse patient populations and clinical settings.

\acks{This work was supported by NSF CAREER award \#2047981, and by a grant to Dr. Elmer from the NIH/NINDS (5R01NS124642). The authors thank the anonymous reviewers for helpful feedback.}

\newpage
\bibliography{ref}

\newpage
\appendix
\section{Notation List} \label{app:notation}
We provide a list of notations used in the paper with their explanation in Table~\ref{tab:notation} and ~\ref{tab:notation2}. 

\begin{table}[htbp]
\centering
\caption{Notations Used in the Competing Risks Setup and Fine and Gray Models}
\label{tab:notation}
\begin{adjustbox} {max width = 1.0\textwidth}
\begin{tabular}{p{2.2cm}p{13cm}} 
  \toprule
    \textbf{Notation} & \textbf{Description} \\
  \midrule
    $[\ell]$ & $\{1,2,\ldots,\ell\}$, for a positive integer $\ell$ \\
    $m$ & Number of competing events \\
    $n$ & Number of training subjects \\
    $i \in [n]$ & Subject index \\
    $k \in [m]$ & Event index \\
    $X_i$ & A random variable representing the feature vector for subject $i$ \\
    $x$ & A realization of a random variable (e.g., a feature vector). \\
    $Y_i$ & The observed time (either the event time or censoring time) for subject $i$ \\
    $D_i$ & The observed event indicator for subject $i$ \\
    $T_{i,k}$ & The latent time of occurrence for event $k$ for subject $i$, with $k\in[m]$ \\
    $C_i$ & The latent censoring time for subject $i$ \\
    
    $F_k(h | x)$ & The cumulative incidence function (CIF) for event $k$ at time horizon $h>0$, given features $x$ \\
    $\lambda_k(h | x)$ & Subdistribution hazard (subhazard) function for event $k$ at time horizon $h$, given features $x$ \\
    $\lambda_{k,0}(h)$ & Baseline subhazard for event $k$ at time horizon $h$\\
    $\phi_k(x;\theta_k)$ & A transformation (e.g., a neural net) of the feature vector $x$ with parameters $\theta_k$, mapping $x$ to a risk score. \\
    $\widehat{\theta}_k$ & The estimated parameters for the risk transformation of event $k$. \\
    $\widehat{\lambda}_{k,0}(h)$ & The estimated baseline subhazard function for event $k$. \\
    $\widehat{\Lambda}_{k,0}(h)$ & The estimated baseline cumulative subhazard for event $k$. \\ 
    $\phi_k(X;\theta_k)$ & Partial subhazard parameterized by $\theta_k$ for feature vector $X$\\
    $L_{\text{IPCW}}(\theta_k)$ & IPCW-adjusted partial likelihood with parameter $\theta_k$ (equation \eqref{eq:ipcw-ll}) \\
    $w_j(h)$ & Weight for subject $j$ at time horizon $h$ used in IPCW adjustment \\
    $\widehat{G}(h)$ & Estimated censoring time distribution at time horizon $h$ \\
    $\widehat{F}_k(h | X)$ & Predicted CIF for event $k$ at horizon $h$, conditioned on feature vector $X$ \\
  \bottomrule
\end{tabular}
\end{adjustbox}
\end{table}

\begin{table}[ht]
\centering
\caption{Notations Used in the Two-Phase Learning}
\label{tab:notation2}
\begin{adjustbox} {max width = 1.0\textwidth}
\begin{tabular}{p{3.2cm}p{13cm}} 
  \toprule
    \textbf{Notation} & \textbf{Description}\\
  \midrule
    $t$ & Prediction time (time after the cardiac arrest onset) \\
    $X_{t}$ & Full feature vector at prediction time $t$ \\
    $X_{t}^{(1)}$ & Phase~1 features: static clinical features as well as the value of $t$ itself (time elapsed since cardiac arrest) \\
    $X_{t}^{(2)}$ & Phase~2 features: time-varying hemodynamic features collected up to time $t$ \\
    $\lambda_{k,t}^{(1)}(h \mid X_t^{(1)})$ & Phase~1 subhazard function for event $k$, based only on $X_t^{(1)}$ \\
    $\phi_{k,t}^{(1)}(X_t^{(1)};\theta_k^{(1)})$ & Phase~1 risk transformation with parameters $\theta_k^{(1)}$ \\
    $\widehat{f}_{k,t}^{(1)}(X_t)$ & Estimated log partial subhazard (risk score) from Phase~1 \\
    $\lambda_{k,t}^{(2)}(h \mid X_t^{(1)},X_t^{(2)})$ & Phase~2 subhazard function for event $k$, incorporating both $X_t^{(1)}$ and $X_t^{(2)}$ \\
    $\phi_{k,t}^{(2)}(X_t^{(1)},X_t^{(2)};\theta_k^{(2)})$ & Phase~2 risk transformation with parameters $\theta_k^{(2)}$ \\
    $\widehat{f}_{k,t}^{(2)}(X_t)$ & Estimated additional risk score from Phase~2 \\
    $I_k(h|X_t)$ & Incremental contribution of Phase~2 features over Phase~1 \\
    $\delta_k(h)$, $\widehat{\delta}_k(h)$ & Thresholding function and its learned estimate for event $k$ at time horizon $h$, used to decide the prediction model \\
    $\widehat{F}_{k,t}^{(1)}(h\mid X_t)$ & Predicted CIF from the Phase~1 model \\
    $\widehat{F}_{k,t}^{(2)}(h\mid X_t)$ & Predicted CIF from the Phase~2 model. \\
    $\widehat{F}_{k,t}(h\mid X_t)$ & Final predicted CIF for event $k$ at time $t$, selected based on the thresholding of $I_k(h|X_t)$ \\
  \bottomrule
\end{tabular}
\end{adjustbox}
\end{table}

\section{More Details on Fine and Gray Competing Risks Modeling} \label{app:fine-gray-model}

\subsection{Estimating Partial Subhazard Function with IPCW-adjusted Partial Likelihood}

Under competing risks, inverse probability of censoring weighting (IPCW) techniques need to be applied to the partial likelihood. The IPCW-adjusted partial likelihood is defined as
\begin{align} \label{eq:ipcw-ll}
  L_{\text{IPCW}}(\theta_k) = \prod_{i=1}^{n} \left[ 
        \frac{\exp\bigl\{\phi_k (X_i; \theta_k)\bigr\}}
        {\sum_{j \in \mathcal{R}_i} w_j(Y_i) \cdot  \exp\bigl\{\phi_k (X_j; \theta_k)\bigr\}}
    \right]^{\mathds{1}\{D_i = k\}},
\end{align}
where $\mathcal{R}_i$ is the (unnatural) risk set of subject $i$ at time $Y_i$ defined as
$$\mathcal{R}_i = \bigl\{j \in [n] ~\big|~ Y_j \geq Y_i \text{ or } (Y_j < Y_i \text{ and } D_j \neq k)\bigr\},$$ 
$w_j(Y_i)$ is the weight for subject $j \in \mathcal{R}_i$ at time $Y_i$ defined as
\begin{align} \label{eq:ipcw-weight}
  w_j(Y_i) = 
  \begin{cases}
    1 & \text{if } Y_j \geq  Y_i, \\
    {\displaystyle \frac{\widehat{G}(Y_i)}{\widehat{G}(Y_j)}} & \text{if }Y_j < Y_i \text{ and } D_j \neq k,
    \end{cases}
\end{align}
and $\widehat{G}(h)$ is an estimate of the censoring time distribution tail function $G(h) = \mathbb{P}(C > h)$.

Then the partial subhazard function (with parameter $\theta_k$) can be estimated by minimizing the negative log-likelihood function, obtained by taking the negative logarithm of the likelihood function defined in equation\eqref{eq:ipcw-ll}, using standard optimization techniques. The same process can be done for each event $k\in[m]$.

\subsection{Estimating Baseline Subhazard with IPCW-adjusted Breslow Method}

Following \cite{fine1999proportional} (also see Chapter 12.5.1 of \citet{collett2023modelling} for reference), once $\phi_k(x;  \widehat \theta_k)$ has been estimated, we can obtain the baseline subhazard function via the Breslow estimator with IPCW adjustment. In the discrete-time setting, this estimator is given by:
\begin{align*}
    \widehat \lambda_{k,0}(h) = \frac{d_k(h)}{\sum_{j \in \mathcal{R}(h)}  w_j(h) \cdot \exp\big\{ \phi_k(x;  \widehat \theta_k) \big\} }
\end{align*}
where $d_k(h)$ is the number of subjects in the training set who experienced event $k$ at time horizon $h$, and $\mathcal{R}(h)$ denotes the ``unnatural'' risk set at time $h$. The weight $w_j(h)$ is the same as defined in equation~\eqref{eq:ipcw-weight}, which is used to adjust for competing risks. Here, the set of distinct time points $h$ corresponds to the unique observed event times in the training set.

Once the baseline subhazard function has been estimated, the baseline cumulative subhazard is computed by summing the estimated baseline subhazard function over the time horizon:
\begin{align*}
    \widehat \Lambda_{k,0}(h) = \sum_{\eta \leq h} \widehat \lambda_{k,0}(\eta)
\end{align*}

\subsection{Modeling with Time-Varying Features at Multiple Time Points} \label{app:model-t}

In our stepwise Fine and Gray model described in Section~\ref{sec:method}, for a certain event $k\in[m]$ of interest, we train a single model that simultaneously learns risk predictions across different time points $t$. 
This is achieved by incorporating feature snapshots from each subject at various times after cardiac arrest. Since subjects have different lengths of observation, we include, for each subject, snapshots of time-varying features (e.g., mean blood pressure and vasopressor dosage) at multiple $t$’s. 
For instance, a snapshot at hour 6 is summarized into a fixed-length vector using summary statistics, and similar snapshots can be taken at hours 7, 8, and so on, until an event occurs or the subject is lost to follow-up. We have the option either to include all available snapshots for each subject or to sample snapshots at predetermined intervals (e.g., every 5 hours). Once these snapshots are extracted for all training subjects, the model is trained on the aggregate of snapshots.

Technically, the Fine and Gray model was originally developed under the assumptions that the data are independent and identically distributed (i.i.d.) and that baseline subhazard estimation shares a common starting time across subjects. Under a dynamic prediction setup, however, we train a single model across various prediction times $t$. This is done by treating different prefixes of a subject’s time series (as more data become available over time) as separate data points, even though these prefixes are not independent. While this strategy enables the model to learn risk predictions at multiple time points, it deviates from the original i.i.d. and common start time assumptions of the Fine and Gray model. We acknowledge this as a limitation of our Stepwise Fine and Gray method.

\section{More Details on the Cohort and the Data} \label{app:data}
For summary statistics of the cohort we use in the paper, see Table~\ref{tab:patient-summary-stats}.

\begin{table}[ht]
  \centering
  \caption{Patient Characteristics Summary Statistics. For most of the covariates, we show the percentage of patients in each event group, except for age and for arrest-to-event time, for which we show the mean. Note that the event group only considers the first event that occurred, i.e., if a patient awakened and then died, then they will only show up in the ``Awakened'' group but not the others.}
  \begin{adjustbox} {max width = 0.9\textwidth}
  \begin{tabular}{lccccc}
    \toprule
    ~ & \multirow{2}{*}{Total} & \multirow{2}{*}{Awakened} & Death, & Death, & Coma \\ 
        ~ & & & not WLST & WLST & (Censoring)\\
    \midrule
      Number of subjects & 2278 & 717 & 672 & 825 & 64 \\ 
      Percentage & 100\% & 31.5\% & 29.5\% & 36.2\% & 2.8\% \\
    \midrule 
      Age (yr) & 57.7 & 57.4 & 55.2 & 59.9 & 58.1 \\ 
      Female & 39.3\% & 34.3\% & 43.5\% & 40.0\% & 43.8\% \\ 
      Out-of-hospital arrest & 83.6\% & 74.5\% & 87.2\% & 88.8\% & 81.3\% \\ 
      Referral from outside facility & 64.2\% & 53.4\% & 69.6\% & 69.7\% & 57.8\% \\ 
      FOUR score - motor & ~ & ~ & ~ & ~ & ~ \\ 
      ~~~~Localizing to pain & 7.5\% & 19.8\% & 3.3\% & 0.5\% & 4.7\% \\ 
      ~~~~Flexion response to pain & 15.2\% & 29.8\% & 5.8\% & 8.8\% & 31.3\% \\ 
      ~~~~Extension response to pain & 2.4\% & 3.6\% & 1.2\% & 2.3\% & 3.1\% \\ 
      ~~~~No response to pain & 51.1\% & 28.7\% & 74.7\% & 52.7\% & 31.3\% \\ 
      ~~~~Myoclonus & 16.2\% & 2.5\% & 10.0\% & 32.6\% & 21.9\% \\ 
      ~~~~Unknown & 7.7\% & 15.5\% & 5.1\% & 3.0\% & 7.8\% \\ 
      Initial rhythm & ~ & ~ & ~ & ~ & ~ \\ 
      ~~~~VT/VF & 27.3\% & 45.9\% & 16.2\% & 19.9\% & 32.8\% \\ 
      ~~~~PEA & 35.4\% & 33.9\% & 36.6\% & 36.0\% & 32.8\% \\ 
      ~~~~Asystole & 30.8\% & 14.8\% & 39.3\% & 37.9\% & 29.7\% \\ 
      ~~~~Unknown & 6.4\% & 5.4\% & 7.9\% & 6.2\% & 4.7\% \\ 
      Arrest-to-event time (hr)  & 106.0 & 83.2 & 85.8 & 107.2 & 556.4 \\ 
      \bottomrule
  \end{tabular} \label{tab:patient-summary-stats} 
  \end{adjustbox}
\end{table}

For blood pressure, non-invasive cuff measurements were recorded when manually taken, whereas invasive mean arterial pressure (MAP) data were continuously monitored via arterial catheter and recorded at approximately hourly intervals. When both types of measurements were available within a given 1-hour window, invasive MAP was prioritized due to its higher reliability (i.e., we ignore any cuff-based mean BP records if there is a MAP record available in the past 1-hour window). Vasopressor administration was tracked for norepinephrine, epinephrine, vasopressin, dopamine, and phenylephrine, with non-zero dosages recorded. 

For Phase 1 of the model, the features comprise static features and the time elapsed since arrest. Including time elapsed allows us to capture the evolution of risk even when dynamic features are not used. In Phase 2, we augment these features with time-varying hemodynamic data to capture additional risk information provided by changes in the patients’ condition.
For a complete list of the features used in both phases, see Table~\ref{tab:covariates}.

\begin{table}[ht]
\centering
\caption{List of Static and Time-Varying Features Used in the Study. Phase 2 features are additional to Phase 1 features, meaning that all Phase 1 features are also included in Phase 2.}
\label{tab:covariates}
\begin{adjustbox} {max width = 0.9\textwidth}
\begin{tabular}{lp{12cm}} 
    \toprule
        \textbf{Feature} & \textbf{Description} \\
        \midrule
        \texttt{age} & Patient age in years \\
        \texttt{female} & Indicator variable for female gender \\
        \texttt{transfer} & Indicator if patient was transferred from another hospital \\
        \texttt{oohca} & Indicator for out-of-hospital cardiac arrest \\
        \texttt{rhythm} & Initial cardiac rhythm observed during arrest; one-hot encoded \\
        \texttt{ca\_type} & Pittsburgh Cardiac Arrest Category\footnote{\url{https://www.emergencymedicine.pitt.edu/patient-care/post-cardiac-arrest-service/pittsburgh-cardiac-arrest-category}}; one-hot encoded \\
        \texttt{FOUR\_Motor} & Motor component score from the FOUR score; one-hot encoded \\
        \texttt{pupils} & Pupillary response (e.g., reactive or non-reactive); one-hot encoded \\
        \texttt{corneals} & Corneal response (e.g., presence/absence of reflex); one-hot encoded \\
        \texttt{t\_since\_arrest\_hr} & Time (in hours) since cardiac arrest \\
    \midrule
        \multicolumn{2}{c}{\textbf{Phase 2: Additional Time-Varying Hemodynamic Features}} \\
        \midrule
        \texttt{mean\_bp} & Mean blood pressure over a 1-hour window \\
        \texttt{min\_bp} & Minimum blood pressure recorded in a 1-hour window \\
        \texttt{max\_bp} & Maximum blood pressure recorded in a 1-hour window \\
        \texttt{bp\_diff} & Change in mean blood pressure from the previous hour (0 if unavailable) \\
        \texttt{DOP\_dose} & Dopamine dosage administered in the 1-hour window \\
        \texttt{EPI\_dose} & Epinephrine dosage administered in the 1-hour window \\
        \texttt{NOR\_dose} & Norepinephrine dosage administered in the 1-hour window \\
        \texttt{VAS\_dose} & Vasopressin dosage administered in the 1-hour window \\
        \texttt{PHE\_dose} & Phenylephrine dosage administered in the 1-hour window \\
        \texttt{cum\_DOP\_dose} & Cumulative dopamine dosage since cardiac arrest \\
        \texttt{cum\_EPI\_dose} & Cumulative epinephrine dosage since cardiac arrest \\
        \texttt{cum\_NOR\_dose} & Cumulative norepinephrine dosage since cardiac arrest \\
        \texttt{cum\_VAS\_dose} & Cumulative vasopressin dosage since cardiac arrest \\
        \texttt{cum\_PHE\_dose} & Cumulative phenylephrine dosage since cardiac arrest \\
    \bottomrule
\end{tabular}
\end{adjustbox}
\end{table}

\begin{figure}[htbp]
   \centering 
   \includegraphics[width=1\linewidth]{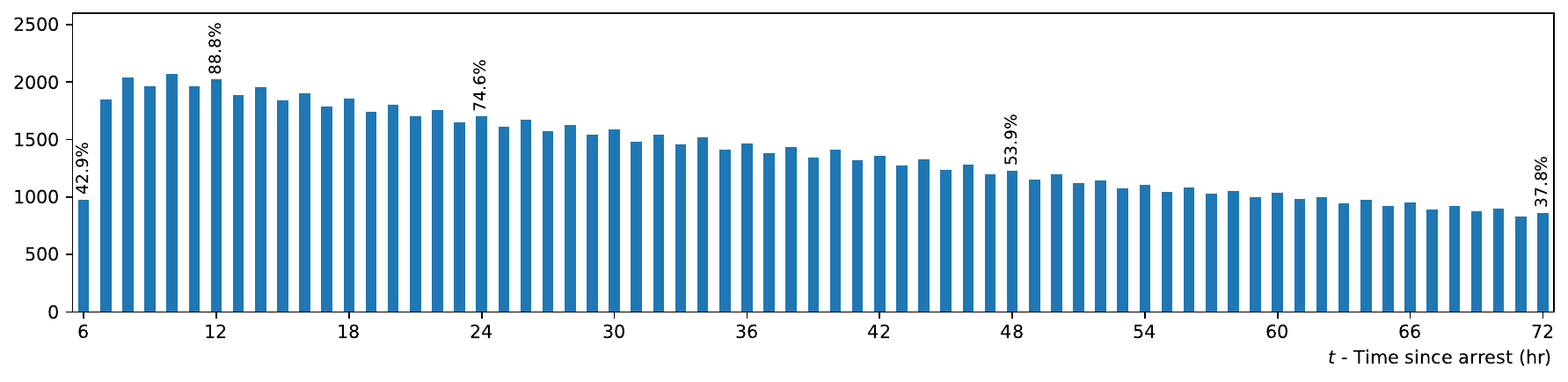} 
   \vspace{-2em}
   \caption{Number of Patients Over Time Since Cardiac Arrest. The bar heights indicate how many patients remain under observation at each hour post-arrest (from a total of 2,278). The percentages on the bar represent the proportion of the total cohort still included at that time. Early gaps primarily reflect delays between the cardiac arrest and the initiation of blood pressure monitoring, whereas later attrition is largely due to patients experiencing their first event (awakening or death). The distribution is long-tailed, and the plot is restricted to a maximum of 72 hours after arrest.} 
   \label{fig:num_patients_at_t} 
\end{figure}

Due to delays in initiating blood pressure monitoring and the occurrence of events (or censoring), blood pressure data are not available from the beginning for all subjects. Besides, if there is a period longer than 5 hours with no blood pressure or vasopressor dosage information, it is likely that the monitoring was interrupted or that the patient was transferred, leading to long gaps in the continuous data needed for reliable risk prediction. Thus, for reliability, we also exclude any data following such long gaps. For details on the number of patients with available data at various time points post-cardiac arrest, see Figure~\ref{fig:num_patients_at_t}.

\section{More Details on the Experiments} \label{app:experiment}
\paragraph{Experiment Configuration}
We parameterize the Fine and Gray models using a neural network architecture. Specifically, the feature encoder consists of two hidden linear layers with 64 and 32 units, respectively, each followed by a ReLU activation and a dropout layer with a dropout rate of 0.2. A final linear layer (with no bias term) produces the output, corresponding to the log partial subhazards. We optimize model parameters using the Adam optimizer with a learning rate chosen from $\{5e-4, 1e-4, 5e-5\}$ and apply a weight decay of 0.001. Training is conducted for up to 1000 iterations, with early stopping triggered if the validation loss does not decrease for 20 consecutive epochs. A batch size of 128 is used, and the best model is selected based on the average CR c-index computed at prediction times $t= 6, 12, 24,$ and $48$ hours on the validation set. For threshold tuning on the incremental contribution from Phase 2, we search for the optimal threshold for each value of horizons $h=$ 24, 48, 72, 120, and 240 hours separately by selecting the threshold that yields the best validation CR c-index. The search grid consists of 100 uniformly sampled values ranging from 0 up to the maximum absolute value of the incremental contribution.

\begin{figure}[htbp]
  \centering
  \includegraphics[width=1\linewidth]{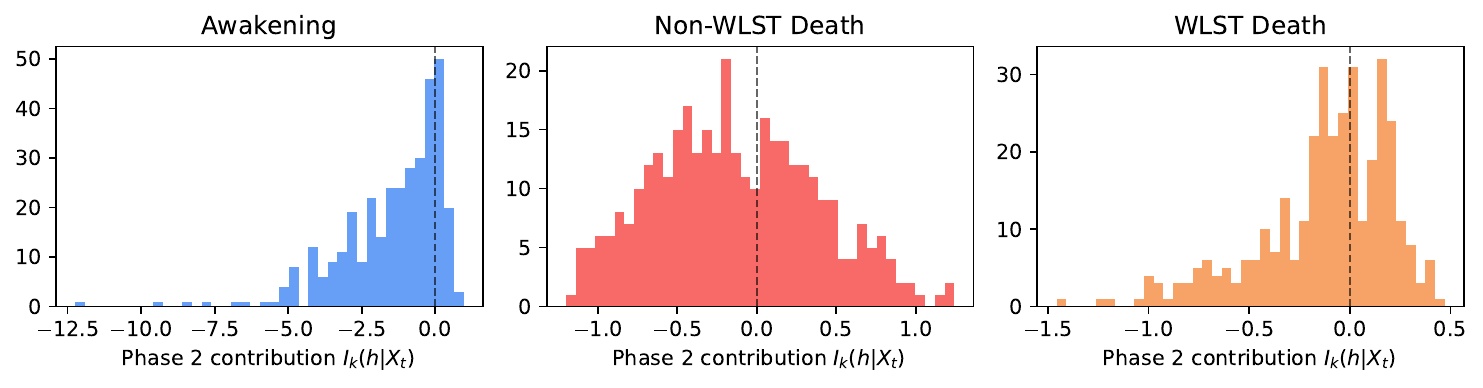}
  \vspace{-2em}
  \caption{Distribution of the Incremental Contribution from Phase 2 for Different Events}
  \label{fig:phase2-contribution}
\end{figure}

\begin{figure}[htbp]
  \centering
  \includegraphics[width=0.5\linewidth]{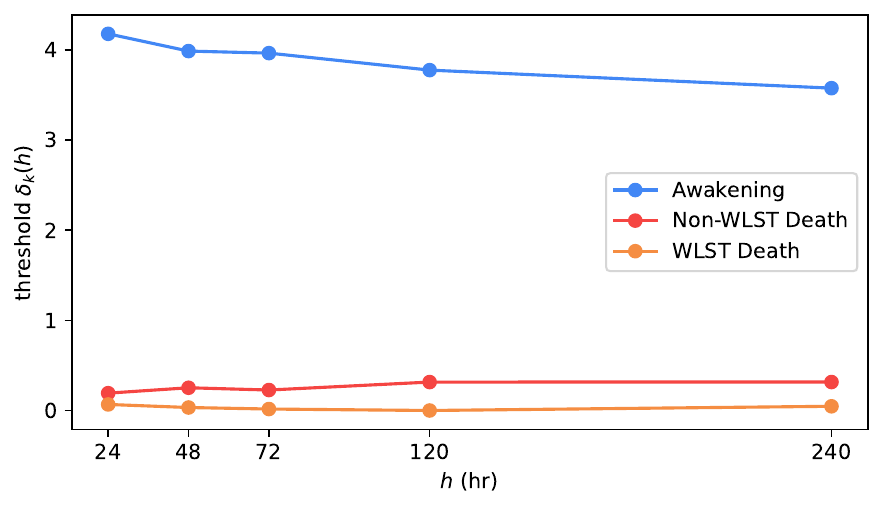}
  \vspace{-1em}
  \caption{Learned Threshold $\delta_k(h)$ of Different Events}
  \label{fig:learned-thres}
\end{figure}

\paragraph{Incremental Contribution and Learned Threshold}
We display the distribution of the incremental contribution from Phase 2 in Figure~\ref{fig:phase2-contribution} and the learned thresholds in Figure~\ref{fig:learned-thres}.

\paragraph{Additional Model Performance at Different $h$ after Tuning Threshold on the Incremental Contribution from Phase 2} For a single trained model, the predicted subhazard values at different time horizons h may lead to varying c-indices. 
In the columns labeled ``Phase 1'' and ``Phase 2'' in Table~\ref{tab:cindex-2nnfg}, the Fine–Gray model’s proportional subhazards assumption guarantees that the c-indices remain identical across different horizons. 
However, after applying the threshold tuning procedure to the incremental contribution from Phase 2, the tuning is performed separately for each $h$ (in our results, we use $h=$~24, 48, 75, 120, and 240 hours), and thus the resulting c-indices can differ across horizons. In Table~\ref{tab:cindex-2nnfg}, for the ``Tuned Threshold'' column, we first compute the mean CR c-index across horizons within one experiment, and then report the overall mean and standard deviation of these values across different experimental repeats. For completeness, we also provide a detailed breakdown of the CR c-index at different horizons in Table~\ref{tab:cindex-diff-h}.

\begin{table}[htbp]
  \centering
  \caption{CR C-index of Two-Phase Fine and Gray with Tuned Threshold at Different Horizons. Prediction time is time since arrest in hour. Values are reported as mean$\pm$standard deviation over 5 experimental repeats.} \label{tab:cindex-diff-h}
  \begin{adjustbox}{max width = 1\textwidth}
    \begin{tabular}{ccccccc}
    \toprule
        \multirow{2}{*}{Event} & Prediction & \multicolumn{5}{c}{Horizon(hr)}  \\ 
        \cmidrule(lr){3-7}
         & Time $t$ (hr) & 24 & 48 & 72 & 120 & 240 \\ 
    \midrule
         \multirow{4}{*}{Awakening} & 6 & 0.9252$\pm$0.0133 & 0.9234$\pm$0.0133 & 0.9240$\pm$0.0138 & 0.9232$\pm$0.0135 & 0.9234$\pm$0.0150 \\ 
        ~ & 12 & 0.9220$\pm$0.0039 & 0.9206$\pm$0.0047 & 0.9203$\pm$0.0047 & 0.9202$\pm$0.0047 & 0.9192$\pm$0.0046 \\ 
        ~ & 24 & 0.9037$\pm$0.0150 & 0.9039$\pm$0.0147 & 0.9040$\pm$0.0148 & 0.9037$\pm$0.0143 & 0.9041$\pm$0.0137 \\ 
        ~ & 48 & 0.8825$\pm$0.0172 & 0.8844$\pm$0.0179 & 0.8844$\pm$0.0179 & 0.8837$\pm$0.0177 & 0.8834$\pm$0.0174 \\ 
    \midrule
        ~ & 6 & 0.8435$\pm$0.0169 & 0.8435$\pm$0.0170 & 0.8433$\pm$0.0170 & 0.8455$\pm$0.0181 & 0.8317$\pm$0.0287 \\ 
        non-WLST & 12 & 0.8496$\pm$0.0244 & 0.8508$\pm$0.0238 & 0.8511$\pm$0.0236 & 0.8504$\pm$0.0240 & 0.8333$\pm$0.0243 \\ 
        death & 24 & 0.8303$\pm$0.0289 & 0.8303$\pm$0.0293 & 0.8304$\pm$0.0291 & 0.8306$\pm$0.0303 & 0.8194$\pm$0.0194 \\ 
        ~ & 48 & 0.8263$\pm$0.0416 & 0.8270$\pm$0.0418 & 0.8270$\pm$0.0413 & 0.8271$\pm$0.0424 & 0.8218$\pm$0.0381 \\ 
    \midrule
        \multirow{4}{*}{WLST} & 6 & 0.8075$\pm$0.0294 & 0.8075$\pm$0.0290 & 0.8077$\pm$0.0294 & 0.8077$\pm$0.0295 & 0.8074$\pm$0.0287 \\ 
        ~ & 12 & 0.8300$\pm$0.0253 & 0.8294$\pm$0.0246 & 0.8297$\pm$0.0248 & 0.8299$\pm$0.0249 & 0.8297$\pm$0.0251 \\ 
        ~ & 24 & 0.8492$\pm$0.0258 & 0.8483$\pm$0.0251 & 0.8487$\pm$0.0254 & 0.8489$\pm$0.0255 & 0.8487$\pm$0.0253 \\ 
        ~ & 48 & 0.8542$\pm$0.0283 & 0.8541$\pm$0.0282 & 0.8547$\pm$0.0276 & 0.8541$\pm$0.0285 & 0.8555$\pm$0.0283 \\ 
    \bottomrule
    \end{tabular}      
  \end{adjustbox}
\end{table}

\paragraph{Model Performance of Other Models}
We report model performance of DeepSurv \citep{faraggi1995neural, katzman2018deepsurv}, DeepHit \citep{lee2018deephit}, Deep Survival Machines~\citep{nagpal2021deep} in Table~\ref{tab:cindex-cr-sota}. Note that DeepSurv does not handle competing risks, and we trained the models by converting other competing events to censoring and training three models for each of the three competing events separately. We can see that our two-phase Fine and Gray model performs as well as or better than other competing risks models, while also providing the additional individual-level variable selection for interpretability.

\begin{table}[htbp]
  \centering
  \caption{CR C-index of the DeepSurv, DeepHit, Deep Survival Machines (DSM). Prediction time is time since arrest in hour. Values are reported as mean$\pm$standard deviation over 5 experimental repeats.}
  \begin{adjustbox}{max width = 1\textwidth}
  \begin{tabular}{cccccc}
    \toprule
    Event & Prediction Time $t$ (hr) & DeepSurv & DeepHit & DSM & Ours\\
    \midrule
      \multirow{4}{*}{Awakening} & 6 & 0.7600$\pm$0.3707 & \bftab{0.9266$\pm$0.0168} & 0.9230$\pm$0.0144 & 0.9238$\pm$0.0138 \\ 
        ~ & 12 & 0.7585$\pm$0.3675 & \bftab{0.9207$\pm$0.0088} & 0.9188$\pm$0.0124 & 0.9205$\pm$0.0044 \\ 
        ~ & 24 & 0.7468$\pm$0.3665 & 0.9107$\pm$0.0097 & \bftab{0.9157$\pm$0.0156} & 0.9039$\pm$0.0145 \\ 
        ~ & 48 & 0.7214$\pm$0.3639 & 0.8834$\pm$0.0202 & \bftab{0.8862$\pm$0.0202} & 0.8837$\pm$0.0176 \\ 
    \midrule
        ~ & 6 & 0.7060$\pm$0.3319 & 0.8278$\pm$0.0132 & 0.8374$\pm$0.0123 & \bftab{0.8415$\pm$0.0191} \\ 
        non-WLST & 12 & 0.7027$\pm$0.3450 & 0.8265$\pm$0.0268 & 0.8378$\pm$0.0277 & \bftab{0.8470$\pm$0.0230} \\ 
        death & 24 & 0.6918$\pm$0.3437 & 0.8255$\pm$0.0257 & \bftab{0.8299$\pm$0.0228} & 0.8282$\pm$0.0260 \\ 
        ~ & 48 & 0.6832$\pm$0.3455 & 0.8220$\pm$0.0351 & 0.8104$\pm$0.0246 & \bftab{0.8258$\pm$0.0382} \\ 
    \midrule
      \multirow{4}{*}{WLST} & 6 & 0.7801$\pm$0.0272 & 0.8055$\pm$0.0181 & 0.7494$\pm$0.0225 & \bftab{0.8076$\pm$0.0271} \\ 
        ~ & 12 & 0.8105$\pm$0.0280 & 0.8274$\pm$0.0196 & 0.7826$\pm$0.0187 & \bftab{0.8297$\pm$0.0232} \\ 
        ~ & 24 & 0.8294$\pm$0.0209 & 0.8455$\pm$0.0232 & 0.8090$\pm$0.0139 & \bftab{0.8488$\pm$0.0236} \\ 
        ~ & 48 & 0.8307$\pm$0.0165 & 0.8428$\pm$0.0314 & 0.8062$\pm$0.0099 & \bftab{0.8545$\pm$0.0262} \\  
      \bottomrule
  \end{tabular} \label{tab:cindex-cr-sota} 
  \end{adjustbox}
\end{table}

\begin{table}[htbp]
  \centering
  \caption{Harrell's C-index of our Two Phase Fine and Gray. Prediction time is time since arrest in hour. However, notice that Harrell's c-index is biased in a competing risks setting. Values are reported as mean$\pm$standard deviation over 5 experimental repeats.}
  \begin{adjustbox}{max width = 0.85\textwidth}
  \begin{tabular}{ccccc}
    \toprule
    Event & Prediction Time $t$ (hr) & Phase 1 & Phase 2 & Tuned Threshold \\
    \midrule
      \multirow{4}{*}{Awakening} & 6 & \bftab{0.7986$\pm$0.0160} & 0.6939$\pm$0.0349 & \bftab{0.7986$\pm$0.0160} \\ 
        ~ & 12 & \bftab{0.7972$\pm$0.0119} & 0.6944$\pm$0.0264 & 0.7956$\pm$0.0094 \\ 
        ~ & 24 & \bftab{0.7984$\pm$0.0137} & 0.6651$\pm$0.0412 & 0.7951$\pm$0.0138 \\ 
        ~ & 48 & 0.7964$\pm$0.0159 & 0.6474$\pm$0.0382 & \bftab{0.7965$\pm$0.0149} \\ 
    \midrule
        ~ & 6 & 0.6534$\pm$0.0212 & 0.7342$\pm$0.0206 & \bftab{0.7370$\pm$0.0186} \\ 
        non-WLST & 12 & 0.6665$\pm$0.0203 & \bftab{0.7744$\pm$0.0306} & 0.7578$\pm$0.0181 \\ 
        death & 24 & 0.6805$\pm$0.0262 & \bftab{0.7524$\pm$0.0369} & 0.7506$\pm$0.0295 \\ 
        ~ & 48 & 0.7335$\pm$0.0463 & 0.7155$\pm$0.0566 & \bftab{0.7653$\pm$0.0460} \\ 
    \midrule
      \multirow{4}{*}{WLST} & 6 & \bftab{0.6790$\pm$0.0300} & 0.5861$\pm$0.0582 & 0.6762$\pm$0.0294 \\ 
        ~ & 12 & \bftab{0.6704$\pm$0.0294} & 0.5695$\pm$0.0396 & 0.6622$\pm$0.0263 \\ 
        ~ & 24 & \bftab{0.6726$\pm$0.0341} & 0.5575$\pm$0.0395 & 0.6624$\pm$0.0322 \\ 
        ~ & 48 & \bftab{0.6481$\pm$0.0339} & 0.5475$\pm$0.0479 & 0.6395$\pm$0.0330 \\ 
      \bottomrule
  \end{tabular} \label{tab:bias-cindex-2nnfg} 
  \end{adjustbox}
\end{table}

\paragraph{Robustness Check Excluding COVID-19 Period Data}\label{app:exclude-covid}
The hospital department we worked with was not substantially affected by the COVID-19 pandemic in terms of resource availability or processes of care. In the included cohort, fewer than 1\% of the arrests were attributed to COVID-19 infection. Even so, as a robustness check, we re-run our experiments excluding data from 2020–2021, which accounted for 23.13\% of the entire cohort.  As shown in Table~\ref{tab:cindex-2nnfg-exclude-covid}, the performance patterns are consistent with those in Table~\ref{tab:cindex-2nnfg}, confirming the prognostic power of hemodynamics.

\begin{table}[htbp]
  \centering
  \caption{CR c-index of the Two-Phase Neural Network Fine and Gray Model (excluding the cohort admitted in 2020-2021). ``Phase~1'' refers to predictions based solely on static features; ``Phase~2'' incorporates time-varying hemodynamic features; ``Tuned~Threshold'' applies a threshold to the Phase 2 contribution to determine its clinical significance. Values are reported as mean $\pm$ standard deviation over 5 experimental repeats.} \label{tab:cindex-2nnfg-exclude-covid} 
  \vspace{-0.5em}
  \begin{adjustbox}{max width = 0.85\textwidth}
  \begin{tabular}{ccccc}
    \toprule
    Event & Prediction Time $t$ (hr) & Phase 1 & Phase 2 & Tuned Threshold \\
    
    \midrule
      \multirow{4}{*}{Awakening} & 6 & 0.9060 $\pm$ 0.0121 & 0.7933 $\pm$ 0.0605 & \bftab{0.9109 $\pm$ 0.0126} \\ 
      ~ & 12 & 0.9082 $\pm$ 0.0115 & 0.8020 $\pm$ 0.0500 & \bftab{0.9095 $\pm$ 0.0114} \\ 
      ~ & 24 & \bftab{0.8920 $\pm$ 0.0107} & 0.7426 $\pm$ 0.0457 & 0.8870 $\pm$ 0.0151 \\ 
      ~ & 48 & \bftab{0.8687 $\pm$ 0.0091} & 0.7118 $\pm$ 0.0645 & 0.8600 $\pm$ 0.0107 \\ 
    \midrule
      ~ & 6 & 0.7612 $\pm$ 0.0257 & 0.8037 $\pm$ 0.0231 & \bftab{0.8270 $\pm$ 0.0209} \\ 
      non-WLST & 12 & 0.7770 $\pm$ 0.0356 & 0.8061 $\pm$ 0.0432 & \bftab{0.8321 $\pm$ 0.0405} \\ 
      Death & 24 & 0.7807 $\pm$ 0.0357 & 0.7843 $\pm$ 0.0254 & \bftab{0.8248 $\pm$ 0.0236} \\ 
      ~ & 48 & 0.8061 $\pm$ 0.0390 & 0.7608 $\pm$ 0.0507 & \bftab{0.8260 $\pm$ 0.0385} \\ 
    \midrule
      \multirow{4}{*}{WLST} & 6 & \bftab{0.8211 $\pm$ 0.0238} & 0.7480 $\pm$ 0.0578 & 0.8205 $\pm$ 0.0233 \\ 
      ~ & 12 & \bftab{0.8546 $\pm$ 0.0248} & 0.7385 $\pm$ 0.0448 & 0.8534 $\pm$ 0.0265 \\ 
      ~ & 24 & \bftab{0.8743 $\pm$ 0.0245} & 0.7525 $\pm$ 0.0391 & 0.8725 $\pm$ 0.0215 \\ 
      ~ & 48 & 0.8744 $\pm$ 0.0492 & 0.7719 $\pm$ 0.0472 & \bftab{0.8752 $\pm$ 0.0480} \\ 
    \bottomrule
  \end{tabular} 
  \end{adjustbox}
  
\end{table}

\section{Additonal Patient Visualization}
\begin{figure}[htbp]
  \centering
  \includegraphics[width=1\linewidth]{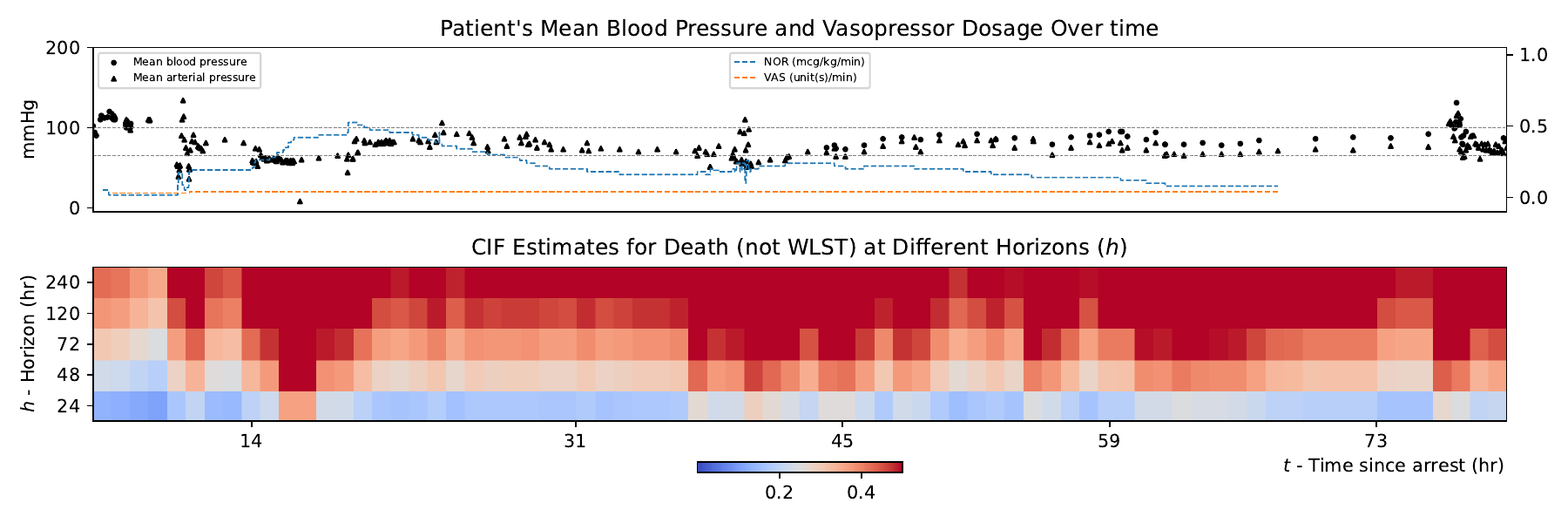}
  \caption{Visualization for a patient who initially presented with poor neurological exam results and ultimately died despite maximal support. Notable fluctuations in the risk estimates correspond to observed variations in the patient’s hemodynamic features.} \label{fig:more-patient-visualization}
\end{figure}

Figure~\ref{fig:more-patient-visualization} represents a patient's feature trajectory and estimated CIF for (non-WLST) death. This patient initially showed no response to pain.

\end{document}